\documentclass[11pt,letterpaper]{article}
\usepackage[letterpaper]{geometry}
\usepackage{acl2012}
\usepackage{times}
\usepackage{latexsym}
\usepackage{amsmath}
\usepackage{multirow}
\usepackage{url}
\usepackage{subcaption}
\makeatletter
\newcommand{\@BIBLABEL}{\@emptybiblabel}
\newcommand{\@emptybiblabel}[1]{}
\makeatother
\usepackage[utf8]{inputenc}
\usepackage{amssymb}
\usepackage{wasysym}
\usepackage{color}
\usepackage{booktabs}
\usepackage{microtype}
\usepackage{tikz-dependency}
\usepackage{hhline}
\usepackage[colorinlistoftodos]{todonotes}

\newcommand{\specificthanks}[1]{\@fnsymbol{#1}}

\title{Cross-Sentence $N$-ary Relation Extraction with Graph LSTMs}

\author{Nanyun Peng\textsuperscript{1}\thanks{~~This research was conducted when the authors were at Microsoft Research.} ~~ Hoifung Poon\textsuperscript{2} ~~ Chris Quirk\textsuperscript{2} ~~ Kristina Toutanova\textsuperscript{3}$^*$ ~~ Wen-tau Yih\textsuperscript{2} \\
\textsuperscript{1} Center for Language and Speech Processing, Computer Science Department \\ 
Johns Hopkins University, Baltimore, MD, USA \\
\textsuperscript{2} Microsoft Research, Redmond, WA, USA \\
\textsuperscript{3} Google Research, Seattle, WA, USA \\
{\tt npeng1@jhu.edu, kristout@google.com} \\
{\tt \{hoifung,chrisq,scottyih\}@microsoft.com}
}

\date{October 2016}

\begin{document}

\maketitle

\begin{abstract}

Past work in relation extraction has focused on binary relations in single sentences. Recent NLP inroads in high-value domains have sparked interest in the more general setting of extracting $n$-ary relations that span multiple sentences. In this paper, we explore a general relation extraction framework based on graph long short-term memory networks (graph LSTMs) that can be easily extended to cross-sentence $n$-ary relation extraction. The graph formulation provides a unified way of exploring different LSTM approaches and incorporating various intra-sentential and inter-sentential dependencies, such as sequential, syntactic, and discourse relations. 
A robust contextual representation is learned for the entities, which serves as input to the relation classifier. This simplifies handling of relations with arbitrary arity, and enables multi-task learning with related relations. We evaluate this framework in two important precision medicine settings, demonstrating its effectiveness with both conventional supervised learning and distant supervision.
Cross-sentence extraction produced larger knowledge bases. and multi-task learning significantly improved extraction accuracy. A thorough analysis of various LSTM approaches yielded useful insight the impact of linguistic analysis on extraction accuracy.
\end{abstract}

\begin{figure*}[t!]
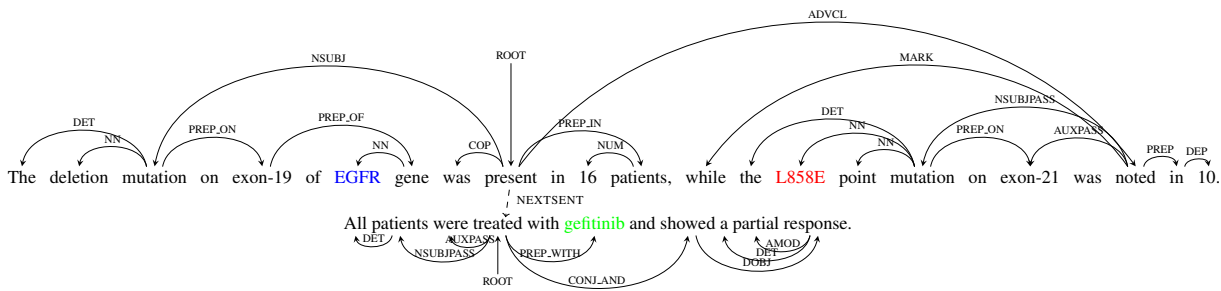

\vspace{-2em}
\footnotesize
\resizebox{\textwidth}{!}{
\begin{dependency}[theme = simple]
   \begin{deptext}[column sep=0.3em]
      The \& deletion \& mutation \& on \& exon-19 \& of \& \color{blue}{EGFR} \& gene \& was \& present \& in \& 16 \& patients, \& while \& the \& \color{red}L858E \& point \& mutation \& on \& exon-21 \& was \& noted \& in \& 10. \\
   \end{deptext}
   \deproot{10}{ROOT}
   \depedge{3}{1}{DET}
   \depedge{3}{2}{NN}
   \depedge{3}{5}{PREP\_ON}
   \depedge{5}{8}{PREP\_OF}
   \depedge{8}{7}{NN}
   \depedge{10}{3}{NSUBJ}
   \depedge{10}{9}{COP}
   \depedge{10}{13}{PREP\_IN}
   \depedge[arc angle=53]{10}{22}{ADVCL}
   \depedge{13}{12}{NUM}
   \depedge{18}{15}{DET}
   \depedge{18}{16}{NN}
   \depedge{18}{17}{NN}
   \depedge{18}{20}{PREP\_ON}
   \depedge[arc angle=55]{22}{14}{MARK}
   \depedge{22}{18}{NSUBJPASS}
   \depedge{22}{20}{AUXPASS}
   \depedge{22}{23}{PREP}
   \depedge{23}{24}{DEP}
   \node (nextsent) [xshift=-6em,yshift=-2.5em] {};
   
   \draw[->] (\wordref{1}{10}) -- (nextsent)[dashed] ;
   \node (nextsentlab) [xshift=-3.5em,yshift=-1.2em] {\tiny\textsc{NEXTSENT}};
   
   \begin{deptext}[column sep=-0.2em]
   \\
   \\
                \&     \&           \&            \&    \&     \&     \&           \&         \&         \&     \&          \&    \&            \&        \&       \&    \&  \              \\
                \&     \&           \&            \&    \&     \&     \&           \&         \&         \&     \&          \&    \&            \&        \&       \&    \&  \              \\
                \&     \&           \&            \&    \&     \&     \&           \&         \&         \&     \&          \&    \&            \&        \&       \&    \&  \              \\
	  All \& patients \& were \& treated \& with \& \color{green}{gefitinib} \& and \& showed  \& a \& partial \& response. \\
   \end{deptext}
   \depedge[edge below]{2}{1}{DET}
   \depedge[edge below]{4}{2}{NSUBJPASS}
   \depedge[edge below]{4}{3}{AUXPASS}
   \depedge[edge below]{4}{6}{PREP\_WITH}
   \depedge[edge below]{4}{8}{CONJ\_AND}
   \deproot[edge unit distance=2ex,edge below]{4}{ROOT}
   \depedge[edge below]{8}{11}{DOBJ}
   \depedge[edge below]{11}{9}{DET}
   \depedge[edge below]{11}{10}{AMOD}
\end{dependency}
}
\vspace{-2.5em}
\caption{
    An example document graph for a pair of sentences expressing a ternary interaction (tumors with L858E mutation in EGFR gene respond to gefitinib treatment).
    For simplicity, we omit edges between adjacent words or representing discourse relations. 
}
\vspace{-1.5em}
\label{fig:doc-graph}
\end{figure*}

\section{Introduction}

Relation extraction has made great strides in newswire and Web domains. 
Recently, there has been increasing interest in applying relation extraction to high-value domains such as biomedicine.
The advent of \$1000 human genome\footnote{\url{http://www.illumina.com/systems/hiseq-x-sequencing-system.html}} heralds the dawn of precision medicine, but progress in personalized cancer treatment has been hindered by the arduous task of interpreting genomic data using prior knowledge.
For example, given a tumor sequence, a molecular tumor board needs to determine which genes and mutations are important, and what drugs are available to treat them.
Already the research literature has a wealth of relevant knowledge, and it is growing at an astonishing rate. PubMed\footnote{\url{https://www.ncbi.nlm.nih.gov/pubmed}}, the online repository of biomedical articles, adds two new papers per minute, or one million each year.
It is thus imperative to advance relation extraction for machine reading.

In the vast literature on relation extraction, past work focused primarily on binary relations in single sentences, limiting the available information. 
Consider the following example: ``{\small \em The deletion mutation on exon-19 of {\bf EGFR} gene was present in 16 patients, while the {\bf L858E} point mutation on exon-21 was noted in 10. All patients were treated with {\bf gefitinib} and showed a partial response.}".
Collectively, the two sentences convey the fact that there is a ternary interaction between the three entities in bold, which is not expressed in either sentence alone. Namely, tumors with {\em L858E} mutation in {\em EGFR} gene can be treated with {\em gefitinib}.
Extracting such knowledge clearly requires moving beyond binary relations and single sentences.

$N$-ary relations and cross-sentence extraction have received relatively little attention in the past. 
Prior work on $n$-ary relation extraction focused on single sentences~\cite{Palmer:2005:PBA:1122624.1122628,mcdonald2005simple} or entity-centric attributes that can be extracted largely independently~\cite{chinchor1998overview,SurdeanuHeng:14}. 
Prior work on cross-sentence extraction often used coreference to gain access to arguments in a different sentence \cite{Gerber:2010:BNS:1858681.1858841,yoshikawa2011coreference}, without truly modeling inter-sentential relational patterns.~(See Section~\ref{sec:related} for a more detailed discussion.)
A notable exception is~\newcite{quirkpoon2017}, which applied distant supervision to general cross-sentence relation extraction, but was limited to binary relations.

In this paper, we explore a general framework for cross-sentence $n$-ary relation extraction, based on graph long short-term memory networks (graph LSTMs).
By adopting the graph formulation, our framework subsumes prior approaches based on chain or tree LSTMs, and can incorporate a rich set of linguistic analyses to aid relation extraction.
Relation classification takes as input the entity representations learned from the entire text, and can be easily extended for arbitrary relation arity $n$.
This approach also facilitates joint learning with kindred relations where the supervision signal is more abundant.

We conducted extensive experiments on two important domains in precision medicine.
In both distant supervision and supervised learning settings, graph LSTMs that encode rich linguistic knowledge outperformed other neural network variants, as well as a well-engineered feature-based classifier.
Multi-task learning with sub-relations led to further improvement.
Syntactic analysis conferred a significant benefit to the performance of graph LSTMs, especially when syntax accuracy was high.

In the molecular tumor board domain, PubMed-scale extraction using distant supervision from a small set of known interactions produced orders of magnitude more knowledge, and cross-sentence extraction tripled the yield compared to single-sentence extraction.
Manual evaluation verified that the accuracy is high despite the lack of annotated examples.

\section{Cross-sentence n-ary relation extraction}

Let $e_1,\cdots,e_m$ be entity mentions in text $T$. Relation extraction can be formulated as a classification problem of determining whether a relation $R$ holds for $e_1,\cdots,e_m$ in $T$.
For example, given a cancer patient with mutation $v$ in gene $g$, a molecular tumor board seeks to find if this type of cancer would respond to drug $d$. 
Literature with such knowledge has been growing rapidly; we can help the tumor board by checking if the $\tt Respond$ relation holds for the $(d,g,v)$ triple.

Traditional relation extraction methods focus on binary relations where all entities occur in the same sentence (i.e., $m=2$ and $T$ is a sentence),
and cannot handle the aforementioned ternary relations.
Moreover, as we focus on more complex relations and $n$ increases, it becomes increasingly rare that the related entities will be contained entirely in a single sentence.
In this paper, we generalize extraction to cross-sentence, $n$-ary relations, where $m>2$ and $T$ can contain multiple sentences.
As will be shown in our experiments section, $n$-ary relations are crucial for high-value domains such as biomedicine, and expanding beyond the sentence boundary enables the extraction of more knowledge.

In the standard binary-relation setting, the dominant approaches are generally defined in terms of the shortest dependency path between the two entities in question, either by deriving rich features from the path or by modeling it using deep neural networks.
Generalizing this paradigm to the $n$-ary setting is challenging, as there are $\binom{n}{2}$  paths.
One apparent solution is inspired by Davidsonian semantics: first, identify a single trigger phrase that signifies the whole relation, then reduce the $n$-ary relation to $n$ binary relations between the trigger and an argument.
However, challenges remain. It is often hard to specify a single trigger, as the relation is manifested by several words, often not contiguous.
Moreover, it is expensive and time-consuming to annotate training examples, especially if triggers are required, as is evident in prior annotation efforts such as GENIA~\cite{kim2009overview}.
The realistic and widely adopted paradigm is to leverage indirect supervision, such as distant supervision~\cite{craven1999constructing,mintz2009distant}, where triggers are not available. 

Additionally, lexical and syntactic patterns signifying the relation will be sparse.
To handle such sparsity, traditional feature-based approaches require extensive engineering and large data.
Unfortunately, this challenge becomes much more severe in cross-sentence extraction when the text spans multiple sentences.

To overcome these challenges, we explore a general relation extraction 
framework 
based on graph LSTMs.
By learning a continuous representation for words and entities, LSTMs can handle sparsity effectively without requiring intense feature engineering.
The graph formulation subsumes prior LSTM approaches based on chains or trees, and can incorporate rich linguistic analyses.

This approach also opens up opportunities for joint learning with related relations.
For example, the $\tt Response$ relation over $d,g,v$ also implies a binary sub-relation over drug $d$ and mutation $v$, with the gene underspecified.
Even with distant supervision, the supervision signal for $n$-ary relations will likely be sparser than their binary sub-relations.
Our approach makes it very easy to use multi-task learning over both the $n$-ary relations and their sub-relations.

\begin{figure}[t!]
  \centering 
  \includegraphics[scale=0.45]{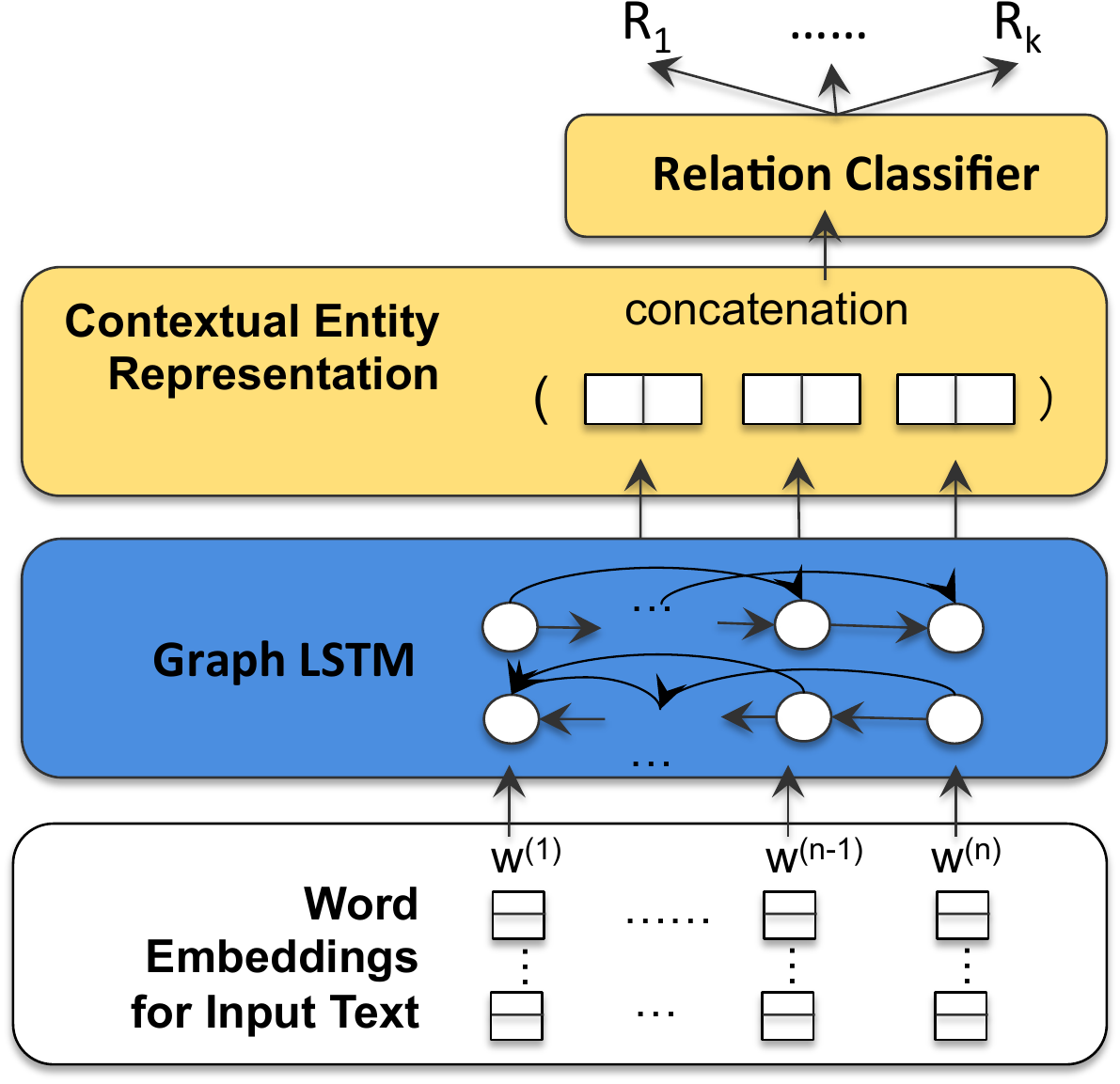}
  \caption{
  A general architecture for cross-sentence $n$-ary relation extraction based on graph LSTMs.
  }
  \label{figure:NN_RE}
\end{figure}

\begin{figure*}[ht!]
\begin{subfigure}[b]{\textwidth}
\begin{dependency}[theme = simple]
   \begin{deptext}[column sep=-0.2em]
	  All \& $\rightleftharpoons$ \& patients \& $\rightleftharpoons$ \& were \& $\rightleftharpoons$ \& treated \& $\rightleftharpoons$ \& with \& $\rightleftharpoons$ \& gefitinib \& $\rightleftharpoons$ \& and \& $\rightleftharpoons$ \& showed \& $\rightleftharpoons$  \& a \& $\rightleftharpoons$ \& partial \& $\rightleftharpoons$ \& response.\\
   \end{deptext}
   \depedge[edge unit distance=1ex]{1}{3}{}
   \depedge[edge unit distance=1ex]{3}{7}{}
   \depedge[edge unit distance=1ex]{5}{7}{}
   \depedge[arc angle=30]{11}{7}{}
   \depedge[arc angle=30]{15}{7}{}
   \depedge[arc angle=30,edge start x offset=6]{21}{15}{}
   \depedge[arc angle=30,edge unit distance=1ex]{17}{21}{}
   \depedge[arc angle=30,edge unit distance=1ex]{19}{21}{}
\end{dependency}
\label{fig:doc-graph-small}
\end{subfigure}
\vspace{-23pt}

\begin{subfigure}[b]{.36\textwidth}
\begin{dependency}[theme = simple]
   \begin{deptext}[column sep=-0.2em]
    $\circ$ \& $\rightarrow$ \& $\circ$ \& $\rightarrow$ \& $\circ$ \& $\rightarrow$ \& $\circ$ \& $\rightarrow$ \& $\circ$ \& $\rightarrow$ \& $\circ$ \& $\rightarrow$ \& $\circ$ \& $\rightarrow$ \& $\circ$ \& $\rightarrow$  \& $\circ$ \& $\rightarrow$ \& $\circ$ \& $\rightarrow$ \& $\circ$\\
   \end{deptext}
   \depedge{1}{3}{}
   \depedge{3}{7}{}
   \depedge{5}{7}{}
   \depedge{17}{21}{}
   \depedge{19}{21}{}
\end{dependency}
\label{fig:forward}
\end{subfigure}
\hspace{6em}
\begin{subfigure}[b]{.36\textwidth}
\begin{dependency}[theme = simple]
   \begin{deptext}[column sep=-0.2em]
   $\circ$ \& $\leftarrow$ \& $\circ$ \& $\leftarrow$ \& $\circ$ \& $\leftarrow$ \& $\circ$ \& $\leftarrow$ \& $\circ$ \& $\leftarrow$ \& $\circ$ \& $\leftarrow$ \& $\circ$ \& $\leftarrow$ \& $\circ$ \& $\leftarrow$  \& $\circ$ \& $\leftarrow$ \& $\circ$ \& $\leftarrow$ \& $\circ$ \\
    \end{deptext}
   \depedge[arc angle=35]{11}{7}{}
   \depedge[arc angle=40]{15}{7}{}
   \depedge[arc angle=40]{21}{15}{}
\end{dependency}
\label{fig:backward}
\end{subfigure}
\vspace{-2em}
\caption{
The graph LSTMs used in this paper. The document graph (top) is partitioned into two directed acyclic graphs (bottom); the graph LSTMs is constructed by a forward pass (Left to Right) followed by a backward pass (Right to Left). Note that information goes from dependency child to parent.
}
\label{fig:doc-graph-dag}
\vspace{-1em}
\end{figure*}

\section{Graph LSTMs}

Learning a continuous representation can be effective for dealing with lexical and syntactic sparsity.
For sequential data such as text, recurrent neural networks (RNNs) are quite popular.
They resemble hidden Markov models (HMMs), except that discrete hidden states are replaced with continuous vectors, and emission and transition probabilities with neural networks.
Conventional RNNs with sigmoid units suffer from gradient diffusion or explosion, making training very difficult~\cite{bengio1994learning,pascanu2013difficulty}.
Long short-term memory (LSTMs)~\cite{hochreiter1997long} combats these problems by using a series of gates (input, forget and output) to avoid amplifying or suppressing gradients during backpropagation.
Consequently, LSTMs are much more effective in capturing long-distance dependencies, and have been applied to a variety of NLP tasks. 
However, most approaches are based on linear chains and only explicitly model the linear context, which ignores a variety of linguistic analyses, such as syntactic and discourse dependencies.

In this section, we propose a general framework that generalizes LSTMs to graphs.
While there is some prior work on learning tree LSTMs \cite{tai2015improved,miwa2016end}, 
to the best of our knowledge, 
graph LSTMs have not been applied to any NLP task yet.
Figure~\ref{figure:NN_RE} shows the architecture of this approach.
The input layer is the word embedding of input text. 
Next is the graph LSTM which learns a contextual representation for each word. 
For the entities in question, their contextual representations
are concatenated and become the input to the relation classifiers.
For a multi-word entity, we simply used the average of its word representations and leave the exploration of more sophisticated aggregation approaches to future work.
The layers are trained jointly with backpropagation.
This framework is agnostic to the choice of classifiers. 
Jointly designing classifiers with graph LSTMs would be interesting future work. 

At the core of the graph LSTM is a {\em document graph} that captures various dependencies among the input words.
By choosing what dependencies to include in the document graph, graph LSTMs naturally subsumes linear-chain or tree LSTMs.

Compared to conventional LSTMs, the graph formulation presents new challenges. 
Due to potential cycles in the graph, a straightforward implementation of backpropagation might require many iterations to reach a fixed point.
Moreover, in the presence of a potentially large number of edge types (adjacent-word, syntactic dependency, etc.), parametrization becomes a key problem.

In the remainder of this section, we first introduce the document graph and show how to conduct backpropagation in graph LSTMs.
We then discuss two strategies for parametrizing the recurrent units.
Finally, we show how to conduct multi-task learning with this framework.

\subsection{Document Graph}

To model various dependencies from linguistic analysis at our disposal, we follow \newcite{quirkpoon2017} and introduce a {\em document graph} to capture intra- and inter-sentential dependencies.
A document graph consists of nodes that represent words and edges that represent various dependencies such as linear context (adjacent words), syntactic dependencies, and discourse relations \cite{lee2013deterministic,xue&al15}.
Figure~\ref{fig:doc-graph} shows the document graph for our running example; this instance suggests that tumors with {\em L858E} mutation in {\em EGFR} gene responds to the drug {\em gefitinib}.

This document graph acts as the backbone upon which a graph LSTM is constructed.
If it contains only edges between adjacent words, we recover linear-chain LSTMs.
Similarly, other prior LSTM approaches can be captured in this framework by restricting edges to those in the shortest dependency path or the parse tree.

\subsection{Backpropagation in Graph LSTMs}

Conventional LSTMs are essentially very deep feed-forward neural networks.
For example, a left-to-right linear LSTM has one hidden vector for each word. This vector is generated by a neural network (recurrent unit) that takes as input the embedding of the given word and the hidden vector of the previous word.
In discriminative learning, these hidden vectors then serve as input for the end classifiers, from which gradients are backpropagated through the whole network.

Generalizing such a strategy to graphs with cycles typically requires unrolling recurrence for a number of steps \cite{scarselli2009graph,li2015gated,liang2016semantic}.
Essentially, a copy of the graph is created for each step that serves as input for the next.
The result is a feed-forward neural network through time, and backpropagation is conducted accordingly.

In principle, we could adopt the same strategy. Effectively, gradients are backpropagated in a manner similar to loopy belief propagation (LBP).
However, this makes learning much more expensive as each update step requires multiple iterations of backpropagation. Moreover, loopy backpropagation could suffer from the same problems encountered to in LBP, such as oscillation or failure to converge.

We observe that dependencies such as coreference and discourse relations are generally sparse, so the backbone of a document graph consists of the linear chain and the syntactic dependency tree.
As in belief propagation, such structures can be leveraged to make backpropagation more efficient by replacing synchronous updates, as in the unrolling strategy, with asynchronous updates, as in linear-chain LSTMs.
This opens up opportunities for a variety of strategies in ordering backpropagation updates.

In this paper, we adopt a simple strategy that performed quite well in preliminary experiments, and leave further exploration to future work.
Specifically, we partition the document graph into two directed acyclic graphs (DAGs). One DAG contains the left-to-right linear chain, as well as other forward-pointing dependencies. The other DAG covers the  right-to-left linear chain and the backward-pointing dependencies.
Figure~\ref{fig:doc-graph-dag} illustrates this strategy. 
Effectively, we partition the original graph into the forward pass (left-to-right), followed by the backward pass (right-to-left), and construct the LSTMs accordingly.
When the document graph only contains linear chain edges, the graph LSTMs is exactly a bi-directional LSTMs (BiLSTMs).

\subsection{The Basic Recurrent Propagation Unit}

A standard LSTM unit consists of an input vector (word embedding), a memory cell and an output vector (contextual representation), as well as several gates. The \emph{input gate} and \emph{output gate} control the information flowing into and out of the cell, whereas the \emph{forget gate} can optionally remove information from the recurrent connection to a precedent unit.

In linear-chain LSTMs, each unit contains only one forget gate, as it has only one direct precedent (i.e., the adjacent-word edge pointing to the previous word). 
In graph LSTMs, however, a unit may have several precedents, including connections to the same word via different edges.
We thus introduce a forget gate for each precedent, similar to the approach taken by \newcite{tai2015improved} for tree LSTMs.

Encoding rich linguistic analysis introduces many distinct edge types besides word adjacency, such as syntactic dependencies, which opens up many possibilities for parametrization.
This was not considered in prior syntax-aware LSTM approaches \cite{tai2015improved,miwa2016end}.
In this paper, we explore two schemes that introduce more fined-grained parameters based on the edge types.

\paragraph{Full Parametrization} 
Our first proposal simply introduces a different set of parameters for each edge type, with computation specified below.

\begin{align}
\small
  i_t &= \sigma(W_i x_t + \sum\nolimits_{j \in P(t)} U^{m(t,j)}_i  h_j + b_i) \nonumber \\ 
  o_t &= \sigma(W_o x_t + \sum\nolimits_{j \in P(t)} U^{m(t,j)}_o h_{j} + b_o) \nonumber \\ 
  \tilde{c_t} &= \tanh(W_c x_t + \sum\nolimits_{j \in P(t)} U^{m(t,j)}_c h_{j} + b_c) \nonumber \\ 
  f_{tj} &= \sigma(W_f x_t + U^{m(t,j)}_f h_j + b_f) \nonumber \\ 
  c_t &= i_t \odot \tilde{c_t} + 
  \sum\nolimits_{j \in P(t)} f_{tj} \odot c_{j} \nonumber \\ 
  h_t &= o_t \odot \tanh(c_t) \nonumber 
\end{align}

As in standard chain LSTMs,
$x_t$ is the input word vector for node $t$, 
$h_t$ is the hidden state vector for node $t$,
$W$'s are the input weight matrices,
and $b$'s are the bias vectors.
$\sigma$, $\tanh$, and $\odot$ represent the sigmoid function, the hyperbolic tangent function, and the Hadamard product (pointwise multiplication), respectively.
The main differences lie in the recurrence terms.
In graph LSTMs, a unit might have multiple predecessors ($P(t)$), for each of which ($j$) there is a forget gate $f_{tj}$, and a typed weight matrix $U^{m(t,j)}$, where $m(t,j)$ signifies the connection type between $t$ and $j$.
The \emph{input} and \emph{output} gates ($i_t, o_t$) depend on all predecessors, 
whereas the forget gate ($f_{tj}$) only depends on the predecessor with which the gate is associated. 
$c_t$ and $\tilde{c_t}$ represent intermediate computation results within the memory cell, which take into account the input and forget gates, and will be combined with output gate to produce the hidden representation $h_t$.

Full parameterization is straightforward, but it requires a large number of parameters when there are many edge types.
For example, there are dozens of syntactic edge types, each corresponding to a Stanford dependency label.
As a result, in our experiments we resort to using only the coarse-grained types: word adjacency, syntactic dependency, etc.
Next, we will consider a more fine-grained approach by learning an edge-type embedding.

\paragraph{Edge-Type Embedding} 
To reduce the number of parameters and leverage potential correlation among fine-grained edge types, we learned a low-dimensional embedding of the edge types, and conducted an outer product of the predecessor's hidden vector and the edge-type embedding to generate a ``typed hidden representation", which is a matrix. 
The new computation is as follows:

\vspace{-5pt}
\begin{align}
\small
  i_t &= \sigma(W_i x_t + \sum\nolimits_{j \in P(t)} U_i \times_T (h_j \otimes e_j) + b_i) \nonumber \\
  f_{tj} &= \sigma(W_f x_t + U_f \times_T (h_j \otimes e_j) + b_f) \nonumber \\
  o_t &= \sigma(W_o x_t + \sum\nolimits_{j \in P(t)} U_o \times_T (h_j \otimes e_{j}) + b_o) \nonumber \\
  \tilde{c_t} &= \tanh(W_c x_t + \sum\nolimits_{j \in P(t)} U_c \times_T (h_j \otimes e_{j}) + b_c) \nonumber \\
  c_t &= i_t \odot \tilde{c_t} + 
  \sum\nolimits_{j \in P(t)} f_{tj} \odot c_{j} \nonumber \\
  h_t &= o_t \odot \tanh(c_t) \nonumber
  \vspace{-5pt}
\end{align}
$U$'s are now $l \times l \times d$ tensors ($l$ is the dimension of the hidden vector and $d$ is the dimension for edge-type embedding), and 
$h_j \otimes e_j$ is a tensor product that produces an $l \times d$ matrix. $\times_T$ denotes 
a tensor dot product defined as $T \times_T A = \sum_d (T_{:,:,d} \cdot A_{:,d})$, which produces an $l$-dimensional vector. The edge-type embedding $e_j$ is jointly trained with the other parameters.

\subsection{Comparison with Prior LSTM Approaches}

The main advantages of a graph formulation are its generality and flexibility. 
As seen in Section 3.1, linear-chain LSTMs are a special case when the document graph is the linear chain of adjacent words.
Similarly, Tree LSTMs \cite{tai2015improved} are a special case when the document graph is the parse tree.

In graph LSTMs, the encoding of linguistic knowledge is factored from the backpropagation strategy (Section 3.2), making it much more flexible, including introducing cycles.
For example, \newcite{miwa2016end} conducted joint entity and binary relation extraction by stacking a LSTM for relation extraction on top of another LSTM for entity recognition. In graph LSTMs, the two can be combined seamlessly using a document graph comprising both the word-adjacency chain and the dependency path between the two entities.

The document graph can also incorporate other linguistic information.
For example, coreference and discourse parsing are intuitively relevant for cross-sentence relation extraction.
Although existing systems have not yet been shown to improve cross-sentence relation extraction \cite{quirkpoon2017}, it remains an important future direction to explore incorporating such analyses, especially after adapting them to the biomedical domains \cite{bell&al16}.

\subsection{Multi-task Learning with Sub-relations}

Multi-task learning has been shown to be beneficial in training neural networks \cite{caruana1998multitask,collobert2008unified,peng2016multi}.
By learning contextual entity representations, our framework makes it straightforward to conduct multi-task learning.
The only change is to add a separate classifier for each related auxiliary relation.
All classifiers share the same graph LSTMs representation learner and word embeddings, and can potentially help each other by pooling their supervision signals. 

In the molecular tumor board domain, we applied this paradigm to joint learning of both the ternary relation (drug-gene-mutation) and its binary sub-relation (drug-mutation).
Experiment results show that this provides significant gains in both tasks.

\section{Implementation Details}

We implemented our methods using the Theano library \cite{2016arXiv160502688short}.
We used logistic regression for our relation classifiers.
Hyper parameters were set based on preliminary experiments on a small development dataset.
Training was done using mini-batched stochastic gradient descent (SGD) 
with batch size 8. 
We used a learning rate of 0.02 and trained for at most 30 epochs, with early stopping based on development data \cite{giles2001overfitting,graves2013speech}. 
The dimension for the hidden vectors in LSTM units was set to 150, and the dimension for the edge-type embedding was set to 3.
The word embeddings were initialized with the publicly available 
100-dimensional GloVe word vectors trained on 6 billion words from Wikipedia
and web text\footnote{\url{http://nlp.stanford.edu/projects/glove/}}~\cite{pennington2014glove}. 
Other model parameters were initialized with random samples drawn uniformly from the range $[-1, 1]$.

In multi-task training, we alternated among all tasks, each time passing through all data for one task\footnote{However, drug-gene pairs have much more data, so we sub-sampled the instances down to the same size as the main $n$-ary relation task.}, and updating the parameters accordingly. This was repeated for 30 epochs.

\section{Domain: Molecular Tumor Boards}

Our main experiments focus on extracting ternary interactions over drugs, genes and mutations, which is important for molecular tumor boards.
A drug-gene-mutation interaction is broadly construed as an association between the drug efficacy and the mutation in the given gene.
There is no annotated dataset for this problem.
However, due to the importance of such knowledge, oncologists have been painstakingly curating known relations from reading papers.
Such a manual approach cannot keep up with the rapid growth of the research literature, and the coverage is generally sparse and not up to date.
However, the curated knowledge can be used for distant supervision.

\subsection{Datasets} 

We obtained biomedical literature from PubMed Central\footnote{\url{http://www.ncbi.nlm.nih.gov/pmc/}}, consisting of approximately one million full-text articles as of 2015.
Note that only a fraction of papers contain knowledge about drug-gene-mutation interactions. Extracting such knowledge from the vast body of biomedical papers is exactly the challenge.
As we will see in later subsections, distant supervision enables us to generate a sizable training set from a small number of manually curated facts, and the learned model was able to extract orders of magnitude more facts.
In future work, we will explore incorporating more known facts for distant supervision and extracting from more full-text articles.

We conducted tokenization, part-of-speech tagging, and syntactic parsing using SPLAT \cite{quirk&al12}, and obtained Stanford dependencies \cite{marneffe&al06} using Stanford CoreNLP \cite{manning&al14}. 
We used the entity taggers from Literome \cite{poon&al14} to identify drug, gene and mutation mentions.

We used the Gene Drug Knowledge Database (GDKD)~\cite{dienstmann&al15} 
and the Clinical Interpretations of Variants In Cancer (CIVIC) 
knowledge base\footnote{\url{http://civic.genome.wustl.edu}} for distant supervision. 
The knowledge bases distinguish fine-grained interaction types, which we do not use in this paper.

\subsection{Distant Supervision}

After identifying drug, gene and mutation mentions in the text, co-occurring triples with known interactions were chosen as positive examples. 
However, unlike the single-sentence setting in standard distant supervision, care must be taken in selecting the candidates.
Since the triples can reside in different sentences, an unrestricted selection of text spans would risk introducing many obviously wrong examples.
We thus followed \newcite{quirkpoon2017} in restricting the candidates to those occurring in a {\em minimal span}, i.e., we retain a candidate only if is no other co-occurrence of the same entities in an overlapping text span with a smaller number of consecutive sentences.
Furthermore, we avoid picking unlikely candidates where the triples are far apart in the document.
Specifically, we considered entity triples within $K$ consecutive sentences, ignoring paragraph boundaries.
$K=1$ corresponds to the baseline of extraction within single sentences. We explored $K\le 3$, which captured a large fraction of candidates without introducing many unlikely ones.

Only 59 distinct drug-gene-mutation triples from the knowledge bases were matched in the text. 
Even from such a small set of unique triples, 
we obtained 3,462 ternary relation instances that can serve as positive examples. 
For multi-task learning, we also considered drug-gene and drug-mutation sub-relations, which yielded 137,469 drug-gene and 3,192 drug-mutation relation instances as positive examples. 

We generate negative examples by randomly sampling co-occurring entity triples without known interactions, subject to the same restrictions above. 
We sampled the same number as positive examples to obtain a balanced dataset\footnote{We will release the dataset at \\ \url{http://hanover.azurewebsites.net}.}.

\begin{table}[t]
    \centering
    \begin{tabular}{lcc}
    \toprule
    \textbf{\small Model}  &  \textbf{\small Single-Sent.} & \textbf{\small Cross-Sent.} \\
        {\small Feature-Based} & 74.7 & 77.7 \\
    \midrule
        {\small CNN}        & 77.5 & 78.1   \\
        {\small BiLSTM}      & 75.3 & 80.1   \\
        {\small Graph LSTM - EMBED} & 76.5  & 80.6  \\
        {\small Graph LSTM - FULL}  & \bf{77.9} & \bf{80.7} \\
    \bottomrule
    \end{tabular}
    \caption{
        Average test accuracy in five-fold cross-validation for 
        drug-gene-mutation ternary interactions.
        Feature-Based used the best performing model in \protect\cite{quirkpoon2017} with features derived from shortest paths between all entity pairs.
    }
    \label{tb:auto_triple}
\end{table}

\begin{table}[t]
    \centering
    \begin{tabular}{lcc}
    \toprule
    \textbf{\small Model}  &  \textbf{\small Single-Sent.} & \textbf{\small Cross-Sent.} \\
    \midrule
        {\small Feature-Based} & 73.9 & 75.2 \\
    \midrule
        {\small CNN}          & 73.0 & 74.9 \\
        {\small BiLSTM}       & 73.9 & 76.0 \\
        {\small BiLSTM-Shortest-Path}  & 70.2 & 71.7  \\
        {\small Tree LSTM}       & \bf{75.9} & 75.9 \\
        {\small Graph LSTM-EMBED} & 74.3 & 76.5 \\
        {\small Graph LSTM-FULL}  & 75.6 & \bf{76.7} \\
    \bottomrule
    \end{tabular}
    \caption{
        Average test accuracy in five-fold cross-validation for 
        drug-mutation binary relations, with an extra baseline using a BiLSTM on the shortest dependency path \protect\cite{xu2015classifying,miwa2016end}.
    }
    \label{tb:auto_drug_mutation}
\end{table}

\subsection{Automatic Evaluation}

To compare the various models in our proposed framework, we conducted five-fold cross-validation, treating the positive and negative examples from distant supervision as gold annotation.
To avoid train-test contamination, all examples from a document were assigned to the same fold.
Since our datasets are balanced by construction, we simply report average test accuracy on held-out folds.
Obviously, the results could be noisy (e.g., entity triples not known to have an interaction might actually have one), but this evaluation is automatic and can quickly evaluate the impact of various design choices.

We evaluated two variants of graph LSTMs: ``Graph LSTM-FULL'' with full parametrization and ``Graph LSTM-EMBED'' with edge-type embedding.
We compared graph LSTMs with three strong baseline systems: a well-engineered feature-based classifier \cite{quirkpoon2017}, a convolutional neural network (CNN) \cite{zeng2014relation,santos2015classifying,wang2016relation}, and 
a bi-directional LSTM (BiLSTM). 
Following \newcite{wang2016relation}, we used input attention for the CNN and a input window size of 5. 
\newcite{quirkpoon2017} only extracted binary relations. We extended it to ternary relations by deriving features for each entity pair (with added annotation to signify the two entity types), and pooling the features from all pairs.

For binary relation extraction, prior syntax-aware approaches are directly applicable. So we also compared with a state-of-the-art tree LSTM system \cite{miwa2016end} and a BiLSTM on the shortest dependency path between the two entities (BiLSTM-Shortest-Path) \cite{xu2015classifying}.

Table~\ref{tb:auto_triple} shows the results for cross-sentence, ternary relation extraction. 
All neural-network based models outperformed the feature-based classifier, illustrating their advantage in handling sparse linguistic patterns without requiring intense feature engineering.
All LSTMs significantly outperformed CNN in the cross-sentence setting, verifying the importance in capturing long-distance dependencies.

The two variants of graph LSTMs perform on par with each other, 
though Graph LSTM-FULL has a small advantage, suggesting that further exploration of parametrization schemes could be beneficial.
In particular, the edge-type embedding might improve by pretraining on unlabeled text with syntactic parses.

Both graph variants significantly outperformed BiLSTMs ($p<0.05$ by McNemar's chi-square test), though the difference is small.
This result is intriguing. In \newcite{quirkpoon2017}, the best system incorporated syntactic dependencies and outperformed the linear-chain variant (Base) by a large margin. So why didn't graph LSTMs make an equally substantial gain by modeling syntactic dependencies?

One reason is that linear-chain LSTMs can already captured some of the long-distance dependencies available in syntactic parses. 
BiLSTMs substantially outperformed the feature-based classifier, even without explicit modeling of syntactic dependencies. The gain cannot be entirely attributed to word embedding as LSTMs also outperformed CNNs.

Another reason is that syntactic parsing is less accurate in the biomedical domain. Parse errors confuse the graph LSM learner, limiting the potential for gain.
In Section 6, we show supporting evidence in a domain when gold parses are available.

We also reported accuracy on instances within single sentences, which exhibited a broadly similar set of trends.
Note that single-sentence and cross-sentence accuracies are not directly comparable, as the test sets are different (one subsumes the other).

We conducted the same experiments on the binary sub-relation between drug-mutation pairs.
Table~\ref{tb:auto_drug_mutation} shows the results, which are similar to the ternary case: 
Graph LSTM-FULL consistently performed the best for both single sentence and cross-sentence instances. 
BiLSTMs on the shortest path substantially underperformed BiLSTMs or graph LSTMs, losing between 4-5 absolute points in accuracy, which could be attributed to the lower parsing quality in the biomedical domain.
Interestingly, the state-of-the-art tree LSTMs \cite{miwa2016end} also underperformed graph LSTMs, even though they encoded essentially the same linguistic structures (word adjacency and syntactic dependency). 
We attributed the gain to the fact that \newcite{miwa2016end} used separate LSTMs for the linear chain and the dependency tree, whereas graph LSTMs learned a single representation for both.

\begin{table}[t]
    \centering
    \begin{tabular}{lcc}
    \toprule
    & \textbf{\small Drug-Gene-Mut.} & \textbf{\small Drug-Mut.} \\
       \midrule
        {\small BiLSTM} & 80.1 & 76.0 \\
        {\small $+$Multi-task} & {\bf 82.4} & 78.1 \\
       \midrule
        {\small Graph LSTM} & 80.7 & 76.7 \\        
        {\small $+$Multi-task} & 82.0 & {\bf 78.5} \\
        \bottomrule
    \end{tabular}
    \caption{
        Multi-task learning improved accuracy for both BiLSTMs and Graph LSTMs.
    }
    \label{tb:multitask}
\end{table}

To evaluate whether joint learning with sub-relations can help, we conducted multi-task learning using Graph LSTM-FULL 
to jointly train extractors for both the ternary interaction and the drug-mutation, drug-gene sub-relations.
Table~\ref{tb:multitask} shows the results. Multi-task learning resulted in a significant gain for both the ternary interaction and the drug-mutation interaction.
Interestingly, the advantage of graph LSTMs over BiLSTMs is reduced with multi-task learning, suggesting that with more supervision signal, even linear-chain LSTMs can learn to capture long-range dependencies that are were made evident by parse features in graph LSTMs.
Note that there are many more instances for drug-gene interaction than others, so we only sampled a subset of comparable size.
Therefore, we do not evaluate the performance gain for drug-gene interaction, as in practice, one would simply learn from all available data, and the sub-sampled results are not competitive.

We included coreference and discourse relations in our document graph. However, we didn't observe any significant gains, similar to the observation in \newcite{quirkpoon2017}. We leave further exploration to future work.

\subsection{PubMed-Scale Extraction}

Our ultimate goal is to extract all knowledge from available text.
We thus retrained our model using the best system from automatic evaluation (i.e., Graph LSTM-FULL) on all available data.
The resulting model was then used to extract relations from all PubMed Central articles.

Table~\ref{tb:recall} shows the number of candidates and extracted interactions.
With as little as 59 unique drug-gene-mutation triples from the two databases\footnote{There are more in the databases, but these are the only ones for which we found matching instances in the text. In future work, we will explore various ways to increase the number, e.g., by matching underspecified drug classes to specific drugs.}, we learned to extract orders of magnitude more unique interactions.
The results also highlight the benefit of cross-sentence extraction, which yields 3 to 5 times more relations than single-sentence extraction.

Table~\ref{tb:gene-drug-var-stats} conducts a similar comparison on unique number of drugs, genes, and mutations. Again, machine reading covers far more unique entities, especially with cross-sentence extraction.

\begin{table}[t]
    \centering
    \begin{tabular}{lrr}
    \toprule
      & \textbf{\small Single-Sent.} & \textbf{\small Cross-Sent.} \\
    \midrule
      {\small Candidates} &   10,873 &   57,033 \\
      $p\ge0.5$  &   1,408 &    4,279 \\
      $p\ge0.9$  &   530 &    1,461 \\
    \midrule
      {\small GDKD + CIVIC} & \multicolumn{2}{c}{59} \\
    \bottomrule
    \end{tabular}
    \caption{
        Numbers of unique drug-gene-mutation interactions
        extracted from PubMed Central articles, compared to that from manually curated KBs used in distant supervision. $p$ signifies output probability.
    }
    \label{tb:recall}
\end{table}

\begin{table}[t]
    \centering
    \begin{tabular}{lrrr}
    \toprule
       & \textbf{Drug} & \textbf{Gene} & \textbf{Mut.} \\
    \midrule
      GDKD + CIVIC & 16 & 12 & 41 \\
    \midrule
      Single-Sent. ($p\ge 0.9$) & 68 & 228 & 221 \\
      Single-Sent. ($p\ge 0.5$) & 93 & 597 & 476 \\
    \midrule
      Cross-Sent. ($p\ge 0.9$) & 103 & 512 & 445 \\
      Cross-Sent. ($p\ge 0.5$) & 144 & 1344 & 1042 \\
    \bottomrule
    \end{tabular}
    \caption{
        Numbers of unique drugs, genes and mutations in extraction from PubMed Central articles, in comparison with that in the manually curated Gene Drug Knowledge Database (GDKD) and  Clinical Interpretations of Variants In Cancer (CIVIC) used for distant supervision. $p$ signifies output probability.
    }
    \label{tb:gene-drug-var-stats}
\end{table}

\subsection{Manual Evaluation}

Our automatic evaluations are useful for comparing competing approaches, but may not reflect the true classifier precision as the labels are noisy.
Therefore, we randomly sampled extracted relation instances and asked three researchers knowledgeable in precision medicine to evaluate their correctness. For each instance, the annotators were presented with the provenance: sentences with the drug, gene, and mutation highlighted. The annotators determined in each case whether this instance implied that the given entities were related.
Note that evaluation does not attempt to identify whether the relationships are true or replicated in follow-up papers; rather, it focuses on whether the relationships are entailed by the text.

We focused our evaluation efforts on the cross-sentence ternary-relation setting.
We considered three probability thresholds: 0.9 for a high-precision but potentially low-recall setting, 0.5, and a random sample of all candidates.
In each case, 150 instances were selected for a total of 450 annotations.
A subset of 150 instances were reviewed by two annotators, and the inter-annotator agreement was 88\%.

Table~\ref{tb:precision} shows that the classifier indeed filters out a large portion of potential candidates, with estimated instance accuracy of 64\% at the threshold of 0.5, and 75\% at 0.9.
Interestingly, LSTMs are effective at screening out many entity mention errors, presumably because they include broad contextual features.

\begin{table}[t]
    \centering
    \begin{tabular}{lrrr}
    \toprule
      &                  & \textbf{Entity} & \textbf{Relation} \\
      & \textbf{Precision} & \textbf{Error} & \textbf{Error} \\
    \midrule
        Random      & 17\% & 36\% & 47\% \\
        $p \ge 0.5$ & 64\% &  7\% & 29\% \\
        $p \ge 0.9$ & 75\% &  1\% & 24\% \\
    \bottomrule
    \end{tabular}
    \caption{
        Sample precision of drug-gene-mutation interactions extracted from PubMed Central articles. $p$ signifies output probability.
    }
    \label{tb:precision}
\end{table}

\section{Domain: Genetic Pathways}

We also conducted experiments on extracting genetic pathway interactions using the GENIA Event Extraction dataset \cite{kim2009overview}.
This dataset contains gold syntactic parses for the sentences, which offered a unique opportunity to investigate the impact of syntactic analysis on graph LSTMs.
It also allowed us to test our framework in supervised learning.

The original shared task evaluated on complex, nested events for nine event types, many of which are unary relations \cite{kim2009overview}.
Following \newcite{poon2015distant}, we focused on gene regulation and reduced it to binary-relation classification for head-to-head comparison.
We followed their experimental protocol by sub-sampling negative examples to be about three times of positive examples.

Since the dataset is not entirely balanced, we reported precision, recall, and F1.
We used our best performing graph LSTM from the previous experiments. 
By default, automatic parses were used in the document graphs, whereas in Graph LSTM (GOLD), gold parses were used instead.
Table~\ref{tb:genia} shows the results. 
Once again, despite the lack of intense feature engineering, linear-chain LSTMs performed on par with the feature-based classifier \cite{poon2015distant}.
Graph LSTMs exhibited a more commanding advantage over linear-chain LSTMs in this domain, substantially outperforming the latter ($p<0.01$ by McNemar's chi-square test).
Most interestingly, graph LSTMs using gold parses significantly outperformed that using automatic parses, suggesting that encoding high-quality analysis is particularly beneficial.

\begin{table}[t]
    \centering
    \begin{tabular}{lccc}
    \toprule
    \textbf{Model}  &  \textbf{Precision} & \textbf{Recall} & \textbf{F1}  \\
    \midrule
        {\small \newcite{poon2015distant}} & 37.5 & 29.9 & 33.2  \\
        {\small BiLSTM} & 37.6 & 29.4 & 33.0 \\
        {\small Graph LSTM} & 41.4 & 30.0 & 34.8 \\
        {\small Graph LSTM (GOLD)} & \bf{43.3} & \bf{30.5} & \bf{35.8} \\
    \bottomrule
    \end{tabular}
    \caption{
        GENIA test results on the binary relation of gene regulation. Graph LSTM (GOLD) used gold syntactic parses in the document graph.
    }
    \label{tb:genia}
\end{table}

\section{Related Work}
\label{sec:related}

Most work on relation extraction has been applied to binary relations of entities in a single sentence.  We first review relevant work on the single-sentence binary relation extraction task, and then review related work on $n$-ary and cross-sentence relation extraction.

\paragraph{Binary relation extraction}  

The traditional feature-based methods rely on carefully designed 
features to learn good models, and often integrate diverse sources of evidence such as word sequences and syntax context 
\cite{kambhatla2004combining,guodong2005exploring,boschee2005automatic,suchanek2006combining,chan2010exploiting,nguyen2014employing}. 
The kernel-based methods design various subsequence or tree kernels 
\cite{mooney2005subsequence,bunescu2005shortest,qian2008exploiting} to 
capture structured information.
Recently, models based on neural networks have advanced the state of the art by automatically learning powerful feature representations \cite{xu2015semantic,zhang2015bidirectional,santos2015classifying,xu2015classifying,xu2016improved}.

Most neural architectures resemble 
Figure~\ref{figure:NN_RE}, where there is a core representation learner 
(blue) that takes word embeddings as input and produces contextual 
entity representations. Such representations are then
taken by relation classifiers to produce the final predictions.
Effectively representing sequences of words, both convolutional~\cite{zeng2014relation,wang2016relation,santos2015classifying} and RNN-based architectures~\cite{zhang2015bidirectional,socher2012semantic,cai2016bidirectional} have been successful. 
Most of these have focused on modeling either the surface word sequences or the hierarchical syntactic structure.  \newcite{miwa2016end} proposed an architecture that benefits from both types of information, using a surface sequence layer, followed by a dependency-tree sequence layer.

\paragraph{$N$-ary relation extraction} Early work on extracting relations between more than two arguments has been done in 
MUC-7,
with a focus on fact/event extraction from news articles~\cite{chinchor1998overview}. Semantic role labeling in the Propbank \cite{Palmer:2005:PBA:1122624.1122628} or FrameNet \cite{baker1998berkeley} style are also instances of $n$-ary relation extraction, with extraction of events expressed in a single sentence. \newcite{mcdonald2005simple} extract $n$-ary relations in a bio-medical domain, by first factoring the $n$-ary relation into pair-wise relations between all entity pairs, and then constructing maximal cliques of related entities. 
Recently, neural models have been applied to semantic role labeling~\cite{fitzgerald-EtAl:2015:EMNLP,roth-lapata:2016:P16-1}.   
These works learned neural representations by effectively decomposing the $n$-ary relation into binary relations between the predicate and each argument, by embedding the dependency path between each pair, or by combining features of the two using a feed-forward network. Although some re-ranking or joint inference models have been employed, the representations of the individual arguments do not influence each other. In contrast, we propose a neural architecture that jointly represents
$n$ entity mentions, taking into account long-distance dependencies and inter-sentential information.

\paragraph{Cross-sentence relation extraction} Several relation extraction tasks have benefited from cross-sentence extraction, including MUC fact and event extraction \cite{swampillai&stevenson11}, record extraction from web pages \cite{wick2006learning}, extraction of facts for biomedical domains \cite{yoshikawa2011coreference}, and extensions of semantic role labeling to cover implicit inter-sentential arguments \cite{Gerber:2010:BNS:1858681.1858841}. These prior works have either relied on explicit co-reference annotation, or on the assumption that the whole document refers to a single coherent event, to simplify the problem and reduce the need for powerful representations of multi-sentential contexts of entity mentions. Recently, cross-sentence relation extraction models have been learned with distant supervision, and used integrated contextual evidence of diverse types without reliance on these assumptions \cite{quirkpoon2017}, but that work focused on binary relations only and explicitly engineered sparse indicator features. 

\paragraph{Relation extraction using distant supervision} Distant supervision has been applied to extraction of binary \cite{mintz2009distant,poon2015distant} and $n$-ary \cite{reschke2014event,li2015imp} relations, traditionally using hand-engineered features. Neural architectures have recently been applied to distantly supervised extraction of binary relations \cite{zeng2015distant}.
Our work is the first to propose a neural architecture for $n$-ary relation extraction, where the representation of a tuple of entities is not decomposable into independent representations of the individual entities or entity pairs, and which integrates diverse information from multi-sentential context. To utilize training data more effectively, we show how multi-task learning for component binary sub-relations can improve performance. Our learned representation combines information sources within a single sentence in a more integrated and generalizable fashion than prior approaches, and can also improve performance on single-sentence binary relation extraction.

\section{Conclusion}
We explore a general framework for cross-sentence $n$-ary relation extraction based on graph LSTMs.
The graph formulation subsumes linear-chain and tree LSTMs and makes it easy to incorporate rich linguistic analysis.
Experiments on biomedical domains showed that extraction beyond the sentence boundary produced far more knowledge, and encoding rich linguistic knowledge provided consistent gain.

While there is much room to improve in both recall and precision, our results indicate that machine reading can already be useful in precision medicine. In particular, automatically extracted facts (Section 5.4) can serve as candidates for manual curation. Instead of scanning millions of articles to curate from scratch, human curators would just quickly vet thousands of extractions. The errors identified by curators offer direct supervision to the machine reading system for continuous improvement. Therefore, the most important goal is to attain high recall and reasonable precision. Our current models are already quite capable.

Future directions include: interactive learning with user feedback; improving discourse modeling in graph LSTMs; exploring other backpropagation strategies; joint learning with entity linking; applications to other domains.

\section*{Acknowledgements}
We thank Daniel Fried and Ming-Wei Chang for useful discussions, as well as the anonymous reviewers and editor-in-chief Mark Johnson for their helpful comments.

\bibliographystyle{acl2012}

\end{document}